%% file: ms.tex
\documentclass[sigconf,nonacm]{acmart}

\usepackage{graphicx}
\usepackage{subfiles}

\usepackage{todonotes}
\usepackage{booktabs}
\PassOptionsToPackage{hyphens}{url}\usepackage{hyperref}

\usepackage{dsfont}
\usepackage{amsmath}

\usepackage{caption}
\usepackage{subcaption}

\usepackage{lipsum}  

\usepackage{xcolor}

\usepackage{pifont}
\newcommand{\cmark}{\ding{51}}%
\newcommand{\xmark}{\ding{55}}%

\AtBeginDocument{%
  \providecommand\BibTeX{{%
    \normalfont B\kern-0.5em{\scshape i\kern-0.25em b}\kern-0.8em\TeX}}}

\begin{document}

\title{HLTCOE at TREC 2023 NeuCLIR Track}

\author{Eugene Yang, Dawn Lawrie, James Mayfield}
\affiliation{%
  \institution{Human Language Technology Center of Excellence, Johns Hopkins University, USA}
  \country{}}
\email{{eugene.yang,lawrie,mayfield}@jhu.edu}

\renewcommand{\shortauthors}{Yang, Lawrie, and Mayfield}

\begin{abstract}
The HLTCOE team applied PLAID, an mT5 reranker, and document translation to the TREC 2023 NeuCLIR track.
For PLAID we included a variety of models and training techniques --
the English model released with ColBERT v2, 
translate-train~(TT), Translate Distill~(TD) and multilingual translate-train~(MTT). 
TT trains a ColBERT model with English queries and passages automatically translated into the document language from the MS-MARCO v1 collection.
This results in three cross-language models for the track, one per language.
MTT creates a single model for all three document languages
by combining the translations of MS-MARCO passages in all three languages
into mixed-language batches.
Thus the model learns about matching queries to passages simultaneously in all languages. 
Distillation uses scores from the mT5 model over non-English translated document pairs to learn how to score query-document pairs.
The team submitted runs to all NeuCLIR tasks: the CLIR and MLIR news task as well as the technical documents task.
\end{abstract}


\maketitle

\section{Introduction}

This year the HLTCOE's primary contribution was in experimentation with multiple ways to fine-tune mPLMs for CLIR and MLIR.
Specifically, we submitted a suite of ColBERT-X models, including Translate-Train~\cite{colbertx} and Translate-Distill~\cite{tdistill} with an effective cross-encoder developed at last tear's NeuCLIR track as the teacher~\cite{unicamp-at-neuclir}.
The HLTCOE also contributed a set of baseline BM25 retrieval runs using Patapsco~\cite{patapsco}
and re-ranked runs with the cross-encoder trained last year for pool enrichment. 
Table~\ref{tab:run-name} summarizes all submitted runs except the BM25 ones. 
In the rest of this paper, we describe our systems, submissions, and observations.

\input{_run_explain}

\section{mT5 Reranker}

At NeuCLIR 2023, the most effective submission was a reranking model using the MonoT5 model with mT5XXL, having 13 billion parameters~\cite{neuclir2022, unicamp-at-neuclir}. 
This submission uses pointwise re-ranking by taking a softmax over the decoding probabilities of two specific tokens (\texttt{\_} and \texttt{\_true} for mT5XXL) as the probability of the document being relevant given the query~\cite{monot5}. 
In this paper, we use the name ``mT5 reranker'' to refer to this specific reranker for convenience. 
The input queries are titles concatenated with descriptions.
For the Chinese CLIR task only we use the reversed order as this has been shown to be more effective in prior works~\cite{unicamp-at-neuclir}.

\section{ColBERT Retrieval with PLAID}
Our systems primarily use PLAID~\cite{plaid}, an implementation of the ColBERT~\cite{colbert} retrieval architecture
that encodes each token as a vector.
Prior work on training CLIR dense retrieval models has demonstrated successes augmenting training queries and passages with translation to match the CLIR target languages~\cite{colbertx}.
In the NeuCLIR 2022 track~\cite{neuclir2022},
our ColBERT-X models trained with Translate-Train
were the most effective end-to-end neural dense retrieval models.\footnote{Given that the authors are also organizers of the track, ColBERT-X runs are marked as manual runs. Although unlikely, performance might have been affected by knowledge that was only accessible to the organizers.} 

All ColBERT-X variants (including PLAID) use the title concatenated with the description as the query. 
We break each document into passages of 180 tokens with a stride of 90. 
For the PLAID index setting, we use one bit for each dimension of the residual vector.
We search for 2500 passages and aggregate passage scores into a document scores using MaxP~\cite{maxp}. 
For comparison, we also submitted a handful of runs using the original ColBERT-X implementation which does not implement the compression technique; such runs are marked using ``ColBERT-X'' as the model in Table~\ref{tab:run-name}.

The runs called ``PLAID\_shard\_by\_date\_1bit\_v1\_tt'' and ``plaid\_-v1\_mtt\_1bit\_date'' ordered the collection by document creation date if known or by download date,
and then created indexes of the three month time windows through the time period of the collection.
Each three-month window was indexed separately in a PLAID index.
Dates were introduced to topics where applicable.
As Table~\ref{tab:date_info} indicates, in some cases, a start date or date range was identified.
Start dates and date ranges were identified manually by looking for topics that contain specific events
and looking for the start or start and end dates of such events in Wikipedia.
These dates were used to identify which segment or segments of the collection would be used to rank documents.

\input{_date_table}

\subsection{Translate-Distill for CLIR using mT5 Reranker}
Distillation is an effective strategy for training a small yet efficient model
that mimics the effectiveness of a larger and computationally more expensive model~\cite{formal2021splade, rocketqa}. 
In our submissions, we explored training a ColBERT-X model to mimic the behavior of the powerful mT5 reranker. 
This training method is known as Translate-Distill and is described in more detail in \citet{tdistill}.

Translate-Distill starts with selecting hard passages for each training query in the MS~MARCO training set. 
We use an \textbf{English} ColBERTv2 model~\cite{colbertv2} to retrieve the top 50 passages for each query from the MS~MARCO training set.
To obtain the teacher scores for each query/passage pair,
the mT5 reranker scores the \textbf{translated} passage
along with the English query. \footnote{\citet{tdistill} finds that this configuration is not the optimal way to obtain scores. Instead English query/passage pairs should be used to obtain scores.}
Finally, we train the ColBERT-X model using these hard passages in the document language,
queries in English, and scores from the mT5 reranker with a KL Divergence loss. 
Translate-Distill is an extension of the ColBERT-X Translate-Train~\cite{colbertx} approach
that distills ranking knowledge from a reranker instead of learning from the contrastive labels. 

\input{_clir_table}

\subsection{Multilingual Translate-Train}

With an eye toward multilingual retrieval (MLIR),
we would like the model to be capable of retrieving documents across a set of languages. 
MTT~\cite{mtt} generalizes TT by translating the training documents into each target language;
this gives the model the ability to retrieve content expressed in any target language.
A key to making MTT successful is to include documents from every target language in each batch.

Unlike training one ColBERT CLIR model with TT for each language pair
(resulting in three models),
we apply the same ColBERT MLIR model trained with MTT to all three language pairs simultaneously.
This produces a single model capable of participation in all of the NeuCLIR tasks.

\section{Sparse Retrieval}

\subsection{Probabilistic Structured Queries}
Probabilistic Structured Queries (PSQ)~\cite{darwish2003probabilistic} is a translation approach that probabilistically matches a token from one language to a distribution of tokens in another. 
This technique can be used to translate queries, documents, or both. 
Prior work~\cite{wang2012matching} has concluded that mapping documents to the query language at indexing time
achieves the best effectiveness while minimizing query latency. 
The resulting documents are bags of probabilistic tokens in the query language. 
They can be indexed as ordinary documents in a sparse retrieval model such as BM25 and HMM with real-valued weights. 

Our submission uses PSQ to translate the documents, and uses a Hidden Markov Model (HMM)~\cite{xu2000cross} for retrieval.

\subsection{Patapsco BM25 Retrieval}

The Patapsco framework~\cite{patapsco}
supports CLIR lexical retrieval through Pyserini~\cite{pyserini}.
Patapsco ensures that language-specific processing is consistent for both queries and documents.
The HLTCOE team submitted BM25 monolingual runs that used human-translated queries to search documents in their native language (QHT),
CLIR runs that used the track-provided machine query translations to search the native documents (QGT),
and runs that used English queries to search the track-provided document translations (DT).
All languages used spaCy~\cite{spacy} for tokenization.
For Russian and English machine translation, spaCy also provided stemming,
while Parsivar~\cite{Mohtaj2018ParsivarAL} was used for Persian stemming. (We did not stem Chinese.)
We explored three query variants: title, description, and title+description. 
We also explored the addition of ten RM3 expansion terms.

\input{_clir_bm25_table}

\section{Results}

\subsection{CLIR Tasks}

The CLIR runs submitted by the HLTCOE are summarized in Tables~\ref{tab:clir-results} (non BM25) and \ref{tab:clir-bm25} (BM25). 
The Translate-Distill models (c2) performed substantially better than the ColBERT-X models trained with Translate-Train (c6). 
The mT5-distilled models also outperformed the monolingual ColBERT model that indexes the translated documents (c4). 
Note that Run (c5) is trained with the PLAID training implementation using both training queries and passages in the original document language,
i.e., a monolingual model in the document language.
That it underperforms the Translate-Train models indicates either that the human translation of the queries is bad (unlikely)
or that training is suboptimal.\footnote{This finding is consistent with the one documented in the working note of our participation of CIRAL@FIRE 2023. }

The student ColBERT-X models (c2) perform on par with their mT5 reranker (c3) teacher, but are much more efficient.
Stacking the teacher reranker on top of the student model (c1) produces the most effective submission across all three languages. 
This gap suggests room for improvement in knowledge distillation training. 

To investigate the stability and compatibility of the compressed document token representation in PLAID retrieval,
run (c7) shards the collection by date using the ColBERT-X model. 
The degradation from the ColBERT-X retrieval (c6) suggests that temporal vocabulary shifts
create incompatibilities in the decompressed document token embeddings used in the final scoring and ranking.
It is also possible that the download date is not a good approximation of the creation date and negatively affects topics for which a date range was introduced, especially an end date since the download date by definition is after the document is created.

\subsection{MLIR Task}
\input{_mlir_table}

The results of our MLIR submissions are summarized in Table~\ref{tab:mlir-results}.
The most effective run among our submissions is the Multilingual Translate-Train (MTT) ColBERT-X model without using the PLAID vector compression (m1). 
Vector compression (m2) provides efficiency in both search time as well as disk space but sacrifices some effectiveness compared to the uncompressed version. 
Both runs using MTT ColBERT-X model outperform the English ColBERTv2 run that indexes the translated documents (m3), which provides a unified platform to compare scores across documents originating in different languages. 

Run (m4) also shares the same MTT ColBERT-X model but sharding the collection by date with date-augmented topics. 
However, such argumentation provides the worst effectiveness.

\subsection{Technical Document Task}

\input{_tech_table}

The technical documents task represents a huge domain shift from news documents and from the MS~MARCO training data,
which in turn leads to different algorithm rankings.
The HLTCOE submitted many of the same variants to this task as it did for the news tasks.
The big differences were the experiments with dense monolingual retrieval represented by (t13), (t14), and (t15) and the addition of BLADE (t9, t10, t11)~\cite{blade}.
BLADE runs were submitted to the news track by the University of Maryland~\cite{umd-2023-participant}.
Table~\ref{tab:tech-results} shows that the mt5 cross-encoder (t1,t2) is superior to the bi-encoder (t3, t6-t12), just as in the news domain.
However, in this case, reranking the BM25 document translation run (t18) outperforms reranking the PLAID translate train run (t7) for nDCG@20
despite the fact that without reranking PLAID translate-train performs better than document translation BM25 in both measures.
The other attribute that stands out is that lexical matching with either PSQ (t4,t5) or BM25 with query translation (t19,t21) outperforms the ColBERT architecture (t6,t7,t8) unless distillation is used (t3). 
This appears to indicate that while MS~MARCO positive and negative examples are insufficient to train the model how to rank technical Chinese abstracts, distillation, which teaches the model how to score documents, is better able to train the model.
PLAID over translated documents is a weaker model than translate-train;
this is also different from the news domain, likely due to machine translation being less effective on technical documents.
BLADE (t9, t10, t11) is also weaker than PSQ, but does outperform BM25 with document translation (t16,t17, t18). Finally, we unintentionally used a faulty training method with JH~Polo (t15),
which led to its dismal performance. 

Turning to the BM25 runs, the poor performance of document translation especially compared to query translation is a big difference between the news domain and the technical documents domain.
In the news domain, document translation has a fairly substantial edge, especially over machine query translation.
For the technical abstracts, document translation is weakest (t16, t17, t18). 

Among the monolingual runs, reranking dense retrieval (t13) is the most effective.
Of particular note, recall at 1000 goes up from the base run (t6) to the reranked run (t1).
This indicates that relevant documents were ranked at levels past 1000,
which the reranking was able to recover.
In addition, BM25 (t19) is nearly as effective as monolingual PLAID (t15),
again indicating that MS~MARCO positive and negative pairs are insufficient to train the model.

\section{Conclusion}

The HLTCOE team participated in all tasks offered in NeuCLIR 2023.
While officially all runs were marked as manual because, as organizers, we created some of the topics,
only the ``date'' runs where we introduce date metadata to each topic had any direct manual input. 
In general, runs reranked with mT5 outperformed end-to-end neural approaches,
which in turn outperformed sparse retrieval models.
Directions for future research include more investigation into the role dates play in topics
as well as exploring other training strategies for MLIR.

\bibliographystyle{ACM-Reference-Format}
\bibliography{sample-base}

\end{document}

%% file: _run_explain.tex
\begin{table*}[]
\caption{HLTCOE runs. Run names in italic are monolingual runs. }\label{tab:run-name}
    \centering
    
\begin{tabular}{cl|cl|c|l}
\toprule
& Run Name & Type &  Model & Query & Description \\
\midrule
& \multicolumn{4}{l}{News Collection with Single Language Documents}  \\
\midrule
(c1) & PLAIDkd-mT5gt-\{td\}                 & Hybrid & PLAID >> mT5 & TD & PLAID with Translate-Distill followed by mt5 reranker \\
(c2) & PLAIDkd-monomt5tt-td                 &  Dense &        PLAID & TD & PLAID with Translate-Distill \\
(c3) & mT5gt-\{td\}                         & Rerank &          mT5 & TD & mT5-XXL reranker \\
(c4) & plaid\_v2\_eng\_1                    &  Dense &        PLAID & TD & English ColBERTv2 with PLAID using DT \\
(c5) & \it{PLAID192mono-td}                 &  Dense &        PLAID & TD & Monolingual PLAID \\   
(c6) & colbertX                             &  Dense &    ColBERT-X & TD & ColBERT-X \\   
(c7) & PLAID\_shard\_by\_date\_1bit\_v1\_tt &  Dense &        PLAID & TD & PLAID with date-sharded indexes \\
(c8) & PSQ-td                               & Sparse &          PSQ & TD & PSQ-HMM \\
(c9) & PSQ-t                                & Sparse &          PSQ &  T & PSQ-HMM \\
\midrule
& \multicolumn{4}{l}{News Collection with Multilingual Documents}  \\
\midrule
(m1) & colbertX                             &  Dense &    ColBERT-X & TD & MTT ColBERT-X \\
(m2) & plaid\_v1\_mtt\_1bit                 &  Dense &        PLAID & TD & MTT ColBERT-X with PLAID using 1 residual bit \\
(m3) & plaid\_v2\_eng\_1                    &  Dense &        PLAID & TD & English ColBERTv2 with PLAID using DT \\
(m4) & plaid\_v1\_mtt\_1bit\_date           &  Dense &        PLAID & TD & MTT PLAID with date-sharded indexes \\
(m5) & PSQraw-td                            & Sparse &          PSQ & TD & Combining CLIR PSQ-HMM scores \\
(m6) & PSQraw-t                             & Sparse &          PSQ &  T & Combining CLIR PSQ-HMM scores \\
\midrule
& Technical Document  \\
\midrule

 (t1) & rerank\_mt5gt\_td                   & Rerank &          mT5 & TD & mT5-XXL reranker using Google translated queries \\
 (t2) & plaid\_tt\_mt5gt\_td                & Hybrid & PLAID >> mT5 & TD & Eng-Zho TT ColBERT-X followed by mT5 reranker \\
 (t3) & plaid\_distilled\_td                &  Dense &        PLAID & TD & Distilled PLAID  \\
 (t4) & psq\_td\_f32                        & Sparse &          PSQ & TD & PSQ-HMM \\
 (t5) & psq\_t\_f32                         & Sparse &          PSQ &  T & PSQ-HMM \\
 (t6) & colbert\_x\_td                      &  Dense &    ColBERT-X & TD & Eng-Zho TT ColBERT-X \\ 
 (t7) & plaid\_tt\_td                       &  Dense &        PLAID & TD & Eng-Zho TT ColBERT-X with PLAID\\ 
 (t8) & plaid\_V2model\_td                  &  Dense &        PLAID & TD & English ColBERT with DT \\ 
 (t9) & blade-d                           & L-Sparse &        BLADE &  T & BLADE \\
(t10) & blade-td                          & L-Sparse &        BLADE & TD & BLADE \\
(t11) & blade-t                           & L-Sparse &        BLADE &  T & BLADE \\
(t12) & plaid\_jhpolo\_td                   &  Dense &        PLAID & TD & Eng-Zho TT ColBERT-X with JH-POLO \\ 
(t13) & \it{plaid\_monozh\_mt5ht\_td}       & Hybrid & PLAID >> mT5 & TD & Chinese ColBERT followed by mT5 reranker \\
(t14) & \it{rerank\_mt5ht\_td}              & Rerank &          mT5 & TD & mT5-XXL reranker using human translated queries \\
(t15) & \it{plaid\_mono\_td}                &  Dense &        PLAID & TD & Chinese ColBERT with PLAID  \\
\bottomrule
\end{tabular}

\end{table*}

%% file: _date_table.tex
\begin{table}[t]
\caption{Topics where dates were introduced as a metadata filter to the topic }\label{tab:date_info}
    \centering
\begin{tabular}{ccc }
\toprule
topic id & start & end \\
\midrule
203 & 3/23/2021 & 3/29/2021 \\
207 & 9/21/2020 &  \\
220 & 4/7/2018 & 4/7/2018 \\
226 & 6/6/2019 & 6/6/2019 \\
231 & 11/30/2018 & 11/30/2018 \\
232 & 1/6/2018 & 1/14/2018 \\
238 & 11/27/2020 &  \\
240 & 12/12/2019 &  \\
244 & 12/11/2017 &  \\
245 & 12/12/2019 &  \\
247 & 3/21/2018 &  \\
249 & 7/12/2019 & 7/12/2019 \\
253 & 4/22/2019 &  \\
255 & 12/2018 &  \\
256 & 3/29/2018 &  \\
257 & 4/15/2019 &  \\
260 & 3/8/2014 &  \\
264 & 7/27/2018 &  \\
265 & 4/1/2020 & 4/1/2020 \\
266 & 2/26/2017 & 2/26/2017 \\
267 & 1/1/2018 & 1/31/2019 \\
273 & 3/11/2018 &  \\
274 & 3/11/2021 &  \\

\bottomrule
\end{tabular}
\end{table}

%% file: _clir_table.tex
\begin{table*}[t]
\caption{CLIR Results}\label{tab:clir-results}
    \centering
    \begin{tabular}{cl|rr|rr|rr}
\toprule
& & \multicolumn{2}{c|}{Persian} & \multicolumn{2}{c|}{Russian} & \multicolumn{2}{c}{Chinese} \\
& & nDCG & R@1k & nDCG & R@1k & nDCG & R@1k \\

\midrule

(c1) & PLAIDkd-mT5gt-\{td\}                 &   \textbf{0.547} &  0.931 &   \textbf{0.540} &  0.916 &   \textbf{0.485} &  0.940 \\
(c2) & PLAIDkd-monomt5tt-td                 &   0.525 &  \textbf{0.936} &   0.500 &  \textbf{0.932} &   0.474 &  \textbf{0.951} \\
\midrule
(c3) & mT5gt-\{td\}                         &   0.519 &  0.878 &   0.529 &  0.838 &   0.483 &  0.863 \\
\midrule
(c4) & plaid\_v2\_eng\_1                    &   0.481 &  0.900 &   0.463 &  0.883 &   0.423 &  0.886 \\
\midrule
(c5) & \it{PLAID192mono-td}                 &   0.466 &  0.822 &   0.424 &  0.830 &   0.404 &  0.867 \\
(c6) & colbertX                             &   0.495 &  0.873 &   0.488 &  0.896 &   0.437 &  0.923 \\
\midrule
(c7) & PLAID\_shard\_by\_date\_1bit\_v1\_tt &   0.456 &  0.797 &   0.467 &  0.865 &   0.407 &  0.875 \\
\midrule
(c8) & PSQ-td                               &   0.423 &  0.850 &   0.366 &  0.785 &   0.324 &  0.840 \\
(c9) & PSQ-t                                &   0.383 &  0.811 &   0.329 &  0.742 &   0.293 &  0.796 \\

\bottomrule
\end{tabular}

\end{table*}

%% file: _clir_bm25_table.tex
\begin{table*}[]
    \caption{BM25 Runs on CLIR Tasks.}\label{tab:clir-bm25}
    \centering
\begin{tabular}{cccc|cc|cc|cc}
\toprule
& &   &   & \multicolumn{2}{c|}{Persian} & \multicolumn{2}{c|}{Russian} & \multicolumn{2}{c}{Chinese} \\
& Translation & Query &    RM3 &   nDCG & R@1k & nDCG & R@1k & nDCG & R@1k \\
\midrule
\multicolumn{4}{c|}{CLIR} & & & & & \\
\midrule

(c10) &  DT &     T & \xmark &   0.379 &  0.838 &   0.332 &  0.721 &   0.346 &  0.790 \\
(c11) &  DT &     T & \cmark &   \textbf{0.397} &  \textbf{0.881} &   0.321 &  0.771 &   0.380 &  0.861 \\
(c12) &  DT &     D & \xmark &   0.346 &  0.782 &   0.307 &  0.725 &   0.340 &  0.766 \\
(c13) &  DT &     D & \cmark &   0.357 &  0.842 &   0.316 &  0.777 &   0.361 &  0.820 \\
(c14) &  DT &    TD & \xmark &   \textbf{0.397} &  0.839 &   0.363 &  0.766 &   0.368 &  0.805 \\
(c15) &  DT &    TD & \cmark &   0.385 &  0.878 &   \textbf{0.355} &  \textbf{0.838} &   \textbf{0.377} &  \textbf{0.863} \\
(c16) & QGT &     T & \xmark &   0.283 &  0.660 &   0.280 &  0.638 &   0.262 &  0.677 \\
(c17) & QGT &     T & \cmark &   0.302 &  0.748 &   0.279 &  0.714 &   0.295 &  0.746 \\
(c18) & QGT &     D & \xmark &   0.303 &  0.702 &   0.276 &  0.667 &   0.269 &  0.661 \\
(c19) & QGT &     D & \cmark &   0.308 &  0.760 &   0.300 &  0.764 &   0.288 &  0.768 \\
(c20) & QGT &    TD & \xmark &   0.328 &  0.743 &   0.349 &  0.732 &   0.319 &  0.747 \\
(c21) & QGT &    TD & \cmark &   0.328 &  0.781 &   0.315 &  0.792 &   0.328 &  0.822 \\

\midrule
\multicolumn{4}{c|}{Monolingual Retrieval} & & & & &\\
\midrule

(c22) & QHT &     T & \xmark &   0.357 &  0.784 &   0.347 &  0.710 &   0.275 &  0.701 \\
(c23) & QHT &     T & \cmark &   0.374 &  0.809 &   0.352 &  0.772 &   0.299 &  0.757 \\
(c24) & QHT &     D & \xmark &   0.347 &  0.759 &   0.357 &  0.703 &   0.287 &  0.641 \\
(c25) & QHT &     D & \cmark &   0.375 &  0.795 &   0.369 &  0.785 &   0.277 &  0.758 \\
(c26) & QHT &    TD & \xmark &   \textbf{0.380} &  \textbf{0.822} &   0.405 &  0.754 &   0.329 &  0.733 \\
(c27) & QHT &    TD & \cmark &   \textbf{0.380} &  0.821 &   \textbf{0.406} &  \textbf{0.815} &   \textbf{0.344} &  \textbf{0.800} \\

\bottomrule
\end{tabular}
\end{table*}

%% file: _mlir_table.tex
\begin{table}[t]
\caption{MLIR Runs}\label{tab:mlir-results}
    \centering
\begin{tabular}{cl|cc }
\toprule
{} & {} &  nDCG@20 &  R@1000 \\
\midrule
 (m1) & colbertX                   &    \textbf{0.362} &   0.771 \\
 (m2) & plaid\_v1\_mtt\_1bit       &    0.359 &   0.780 \\
 (m3) & plaid\_v2\_eng\_1          &    0.355 &   \textbf{0.804} \\
 (m4) & plaid\_v1\_mtt\_1bit\_date &    0.335 &   0.720 \\
\midrule
 (m5) & PSQraw-td                  &    0.295 &   0.693 \\
 (m6) & PSQraw-t                   &    0.256 &   0.667 \\

\midrule
& BM25 Runs with DT & \\
\midrule

 (m7) & T w/ RM3                   &    0.288 &   0.750 \\
 (m8) & T w/o RM3                  &    0.261 &   0.682 \\
\midrule
 (m9) & D w/ RM3                   &    0.282 &   0.712 \\
(m10) & D w/o RM3                  &    0.251 &   0.662 \\
\midrule
(m11) & TD w/ RM3                  &    0.306 &   0.764 \\
(m12) & TD w/o RM3                 &    0.290 &   0.721 \\

\bottomrule
\end{tabular}

\end{table}

%% file: _tech_table.tex
\begin{table}[t]
\caption{Technical Document Track Results. Italicized names indicate monolingual runs.}\label{tab:tech-results}
    \centering

\begin{tabular}{cl|cc}
\toprule
& {} &  nDCG@20 &  R@1000 \\
\midrule
 (t1) &  rerank\_mt5gt\_td        &    \textbf{0.394} &   0.656 \\
 (t2) &  plaid\_tt\_mt5gt\_td     &    0.383 &   0.755 \\
 (t3) &  plaid\_distilled\_td     &    0.360 &   \textbf{0.824} \\
 (t4) &  psq\_td\_f32             &    0.314 &   0.768 \\
 (t5) &  psq\_t\_f32              &    0.310 &   0.760 \\
 (t6) &  colbert\_x\_td           &    0.287 &   0.693 \\
 (t7) &  plaid\_tt\_td            &    0.279 &   0.717 \\
 (t8) &  plaid\_V2model\_td       &    0.273 &   0.749 \\
 (t9) &  blade-d                  &    0.260 &   0.716 \\
(t10) &  blade-td                 &    0.246 &   0.717 \\
(t11) &  blade-t                  &    0.240 &   0.724 \\
(t12) &  plaid\_jhpolo\_td        &    0.072 &   0.212 \\
\midrule
(t13) &  \it{plaid\_monozh\_mt5ht\_td} &    \textbf{0.410} &   \textbf{0.812} \\
(t14) &  \it{rerank\_mt5ht\_td}   &    0.378 &   0.722 \\
(t15) &  \it{plaid\_mono\_td}     &    0.359 &   0.758 \\
\midrule
& BM25 + RM3 Runs & \\
\midrule
(t16) &   DT +  T                 &    0.222 &   0.660 \\
(t17) &   DT +  D                 &    0.189 &   0.616 \\
(t18) &   DT +  TD                &    0.210 &   0.656 \\
(t19) &  QGT +  T                 &    \textbf{0.322} &   0.793 \\
(t20) &  QGT +  D                &    0.290 &   0.755 \\
(t21) &  QGT +  TD                &    0.314 &   \textbf{0.814} \\
\midrule
(t22) &  \it{QHT +  T }                &    \textbf{0.350} &   0.812 \\
(t23) &  \it{QHT +  D }                &    0.276 &   0.722 \\
(t24) &  \it{QHT +  TD }               &    0.331 &   \textbf{0.813} \\
\bottomrule
\end{tabular}

\end{table}